\documentclass[a4paper]{article}

\usepackage{INTERSPEECH2022}
\usepackage{subcaption}
\usepackage{float}
\usepackage{xcolor}

\newcommand{\softmax}[1]{\sfx\left(#1\right)}
\DeclareMathOperator*{\sfx}{softmax}

\newcommand\blfootnote[1]{%
  \begingroup
  \renewcommand\thefootnote{}\footnote{#1}%
  \addtocounter{footnote}{-1}%
  \endgroup
}

\title{On-demand compute reduction with stochastic wav2vec 2.0}
\name{Apoorv Vyas $^{1,2,*}$\thanks{* work done during internship at Meta AI},
Wei-Ning Hsu $^3$, Michael Auli $^3$, Alexei Baevski $^3$ }
\address{
  $^{1}$Idiap Research Institute, Switzerland \\
  $^{2}$Ecole Polytechnique F\'ed\'erale de Lausanne, Switzerland\\
  $^{3}$Meta AI 
}
\email{avyas@idiap.ch, \{wnhsu,michaelauli,abaevski\}@fb.com}

\begin{document}

\maketitle

\begin{abstract}

    Squeeze and Efficient Wav2vec (SEW) is a recently proposed architecture
    \cite{felix2021performance}  that squeezes the input to the transformer
    encoder for compute efficient pre-training and inference with wav2vec 2.0 (W2V2) models.  In this work, we 
    propose stochastic compression for on-demand compute reduction for W2V2 models.
    As opposed to using a fixed squeeze factor, we sample it uniformly during training. 
    We further introduce query and
    key-value pooling mechanisms that can be applied to each
    transformer layer for further compression. Our results for models
    pre-trained on $960$h Librispeech dataset and fine-tuned on $10$h of transcribed data
    show that using the same stochastic model, we get a smooth trade-off
    between word error rate (WER) and inference time with only marginal WER degradation
    compared to the W2V2 and SEW models trained for a specific setting. We further show that
    we can fine-tune the same stochastically pre-trained model to a specific
    configuration to recover the WER difference resulting in significant computational
    savings on pre-training models from scratch.
\end{abstract}

\noindent\textbf{Index Terms}: self-supervision, speech recognition, wav2vec2

\section{Introduction}
\blfootnote{Preprint. Submitted to Interspeech 2022.}Self-supervised pretraining
\cite{baevski2020wav2vec,chung2019unsupervised,liu2020tera,oord2019representation,schneider2019wav2vec,wang2020speechbert,hsu2021hubert,baevski2022data2vec}
is a powerful technique that enables learning useful representations from
untranscribed audio. Pre-trained models can subsequently be fine-tuned on
a downstream task with supervised data such as automatic speech recognition.

Currently, wav2vec 2.0 (W2V2) \cite{baevski2020wav2vec} is one of the most well performing
self-supervised learning approaches. W2V2 model comprises
a convolutional front-end and a transformer encoder that learns
representations from raw audio data using contrastive learning. However, the
quadratic complexity of self-attention computation together with the long
sequences encountered in speech results in high computational requirements for
training as well as high latency for inference.


%
 
Recently, \cite{felix2021performance} improves the W2V2 architecture for efficient training and inference. The proposed \emph{SEW} architecture modifies the convolutional feature extractor to improve latency. 
They also introduce squeezing
mechanism that downsamples the convolutional front-end output from $50$Hz to $25$Hz. This
reduces the computation for the transformer encoder. 
The output of the encoder is upsampled with a linear layer to produce output at $50$Hz.
One downside to SEW is the WER degradation for sensitive applications.

We propose to overcome this limitation with stochastic pre-training
and fine-tuning for W2V2 models.  Deep neural networks with stochastic depth
using layerdrop and block drops have previously been explored in
\cite{wu2018blockdrop,fan2020reducing,gao2016deep,liu2018rethinking} for efficient inference. In this work,
we introduce stochastic compression for on-demand compute reduction for W2V2
model. Stochastic compression enables training a single model that can support a number of
operating points with a smooth trade-off between WER and inference latency.

As opposed to training with a fixed squeezing factor, at each iteration,
we sample a squeezing factor $S$ from $\{1 \dots S_f\}$. Given the squeezing
factor $S$, we compress the input to the transformer encoder by this factor.
Similar to SEW, the output of the transformer encoder is upsampled to the original
sequence length using a linear layer. In addition to this, for each transformer
layer, we also introduce a stochastic pooling mechanism that could be
independently applied to queries and keys-values in a decoupled fashion.
In contrast to the squeezing mechanism, we do this without introducing any
additional learnable parameters.

We show that stochastic pre-training and fine-tuning provides a smooth
trade-off between WER and inference time with only marginal performance
degradation compared to the non-stochastic variants. 
We further show
that by fine-tuning the same stochastically pre-trained model to a specific
configuration (operating point), we can get the same accuracy as the
corresponding non-stochastic model. This removes the need for pre-training multiple models,
resulting in significant computational savings.


\section{Our Method}
\label{sec:our_method}
We start with a brief background on the Squeeze and Efficient
W2V2 (SEW) model introduced in \cite{felix2021performance}. We then
discuss the proposed query and key-value mean-pooling to further reduce the
context length for the transformer layer.  We finally present the  proposed
stochastic compression for W2V2 training.

\subsection{Squeeze and Efficient Wav2vec (SEW)}
Wu et. al. \cite{felix2021performance} propose two main architectural changes
to the W2V2 architecture. First, they introduce compact wave feature
extractor (WFE-C) that replaces the original W2V2 convolutional feature
extractor (WFE-O). WFE-O uses the same number of channels in all layers of its 
convolutional extractor. WFE-C starts with a small number of channels \emph{c}
and doubles the channel when the sequence length is downsampled by 4 times.
WFE-C distributes the forward and backward pass computation more
evenly across layers resulting in a similar WER as WFE-O while being
much faster. In this work, we always use WFE-C-c$64$-I$1$ as the compact feature
extractor. We refer the readers to section $4.4$ of \cite{felix2021performance}
for more details. 

\begin{figure}
    \centering
    \includegraphics[width=.7\columnwidth]{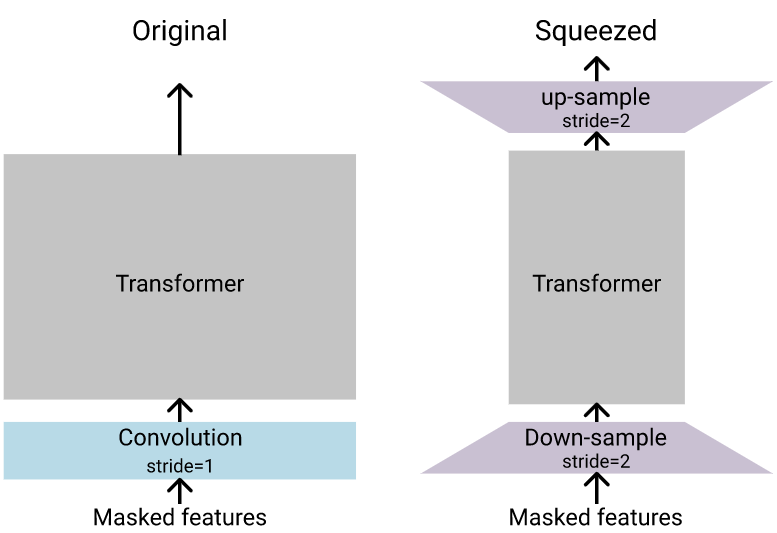}
    \caption{
        Comparing original W2V2 and SEW model architecture with a squeezing
        factor of 2. Figure taken from \cite{felix2021performance}
    }
    \label{fig:sew}
\end{figure}

Next, they introduce \textbf{Squeezed Context Networks} to reduce the length of
input sequence to the transformer encoder. As shown in Fig. \ref{fig:sew}, they
introduce a \emph{squeezing} mechanism via a downsampling layer at the output of the convolutional 
encoder to reduce the output rate from $50$Hz to $25$Hz. The downsampled sequence
reduces the memory and computational time for the transformer
encoder. The upsampling layer at the output of transformer encoder
produces outputs at $50$Hz for contrastive loss computation.

\subsection{Basic Notations}

We start by introducing  basic definitions that we will later use to define
query and key-value mean pooled attention. Let us denote the queries as $Q \in
\mathbb{R}^{N \times D_k}$, keys as $K \in \mathbb{R}^{S \times D_k}$, and
values as $V \in \mathbb{R}^{S \times D_v}$, where $N$ and $S$ denotes the
query and key sequence lengths respectively. $D_k$ and $D_v$ denote the
embedding dimensions for keys and values respectively. The attention output
$V^\prime \in \mathbb{R}^{N \times D_v}$ is given as follows:

\begin{equation}
    A(Q, K, V) = V^\prime = \softmax{\frac{Q K^T}{\sqrt{D_k}}} V \label{eq:attention}.
\end{equation}

Let us also define the mean-pooling or downsampling operator $D(X, S_p)$ that takes as input a sequence
$X \in \mathbb{R}^{N \times D_x}$ and a pooling factor $S_p$ to output another
sequence $X^p \in \mathbb{R}^{N_p \times D_x}$ where $N_p = \lceil \frac{N}{S_p}
\rceil$. Here $\lceil a \rceil$ denotes the \emph{ceil} operation. Note that we may need to pad the signal appropriately. The i-$th$ output $X^p_i$ is given by:


\begin{equation}
\begin{aligned}
    X^p_i = \sum_{j=1}^{S_p} \frac{X_{(i*S_p) + j}}{S_p} \quad \forall i \in \{1,\dots,N_p\}.\\
\end{aligned}
\end{equation}

Finally, let us define the upsampling operator $U(X^p, S_u)$ that takes as input a sequence
$X^p \in \mathbb{R}^{N_p \times D_x}$ and an upsampling factor $S_u$ to output another
sequence $X \in \mathbb{R}^{N \times D_x}$ where $N = (N_p * S_u)$. The i-$th$ output $X_i$ is given by:

\begin{equation}
\begin{aligned}
     X_i = X^p_{\lfloor \frac{i}{S_u} \rfloor} \quad \forall i \in \{1,\dots,N\},\\
\end{aligned}
\end{equation}
where $\lfloor a \rfloor$ denotes the \emph{floor} operation. 


\subsection{Query and Key-Value Mean Pooled Attention}
The upsampling layer  in the SEW architecture introduces some additional
parameters to the W2V2 model. In contrast to this, we introduce query
and key-value mean pooling during self-attention computation which can be applied
independently to each layer with no additional parameters.  This allows for finer control over the compression at each transformer layer. 

Let us denote the query pooling and key-value pooling factors as $S_q$ and
$S_k$ respectively. Given $S_q$ and $S_k$, we first compute the mean pooled
queries, keys, and values as follows:

\begin{align}
   Q_p &= D(Q, S_q), \\ 
   K_p &= D(K, S_k), \\ 
   V_p &= D(V, S_k).
\end{align}

The output $V^\prime \in \mathbb{R}^{N \times D_v}$ for the pooled attention is:
\begin{align}
   V^\prime_p &= A(Q_p, K_p, V_p), \\
   V^\prime &= U(V^\prime_p, S_q).
\end{align}

\subsection{Stochastic Compression}

In contrast to SEW models, where the squeezing factor remains fixed during
pre-training and fine-tuning, at each iteration, we sample the squeezing
factor uniformly from the set $\{1,\dots, S_f\}$. Similarly, for each
transformer layer we also sample the query and key-value mean pooling factors
from the sets $\{1,\dots,S_q\}$ and $\{1,\dots,S_k\}$ respectively. 

\section{Experiments}
\label{sec:experiments}
\subsection{Models}
We conduct experiments with W2V2, SEW, and our proposed stochastically
compressed (St-SEW) models with $12$ and $24$ encoder layers.  For all models, we use WFE-C-c$64$-l$1$ as the feature extractor.
In Table \ref{tab:hyper-params}, we describe the main
hyper-parameters for different classes of models. Note that we can view 
W2V2 and SEW models as special cases of the stochastic models with a specific
setting of squeezing and pooling factors.

\begin{table}
    \caption{
        Model hyper-parameters and pre-training times in hours  for base (B) and large (L) models. $S_f$ refers
        to possible squeeze factors, $S_q$ and $S_k$ refer to the possible
        query and key-value pooling factors. E and D refer to the model dimension and the number of layers (depth) in transformer respectively.
        We estimate the pre-training times (PT) for $400$K steps for models trained on $8$ NVIDIA V100 GPUs.
    }
    \centering
    \label{tab:hyper-params}
    \begin{tabular}{l c c c c c c}
        \toprule
        Model     & $S_f$   & $S_k$   & $S_q$   & E     & D  & PT (hr) \\
        W2V2-B     & 1       & 1       & 1      & 768  & 12 & 117 \\
        W2V2-L     & 1       & 1       & 1      & 1024 & 24 & 172 \footnotemark \\
        \midrule
        SEW-B     & 2       & 1       & 1       & 768  & 12  & 83 \\
        SEW-L     & 2       & 1       & 1       & 1024 & 24  & 115 \\
        \midrule
        St-SEW-B  & \{1,2\}   & \{1,2\}   & \{1,2\}   & 768  & 12 & 100 \\
        St-SEW-L  & \{1,2\}   & \{1,2\}   & \{1,2\}   & 1024 & 24 & 140 \\
        \bottomrule
    \end{tabular}
\end{table}
\footnotetext[1]{Estimated time. The model is unstable and pre-training diverged.}

\subsection{Pre-training}
We pre-train models on $960$h of Librispeech \cite{panayotov2015librispeech}
dataset.  We  use  the  same hyperparameters as W2V2 base
\cite{baevski2020wav2vec}. To reduce the computational requirements, our
models are trained with 8 NVIDIA V100 GPUs using Fairseq \cite{ott2019fairseq} and PyTorch \cite{paszke2019pytorch}. We double the maximum tokens per batch and set gradient accumulation steps to 4 to simulate 64 GPUs
as used in \cite{baevski2020wav2vec}.

\subsection{Fine-tuning}

We add a linear layer to the top of the transformer encoder and fine-tune the
model  for $20$K steps using the CTC objective \cite{graves2006ctc} on the $10$h
subset of the Librispeech dataset.  We use the \emph{dev-other} for model selection
during fine-tuning. For stochastically pre-trained models, we consider the
following two strategies for fine-tuning: \\

\noindent\textbf{Stochastic Fine-tuning}
In this setting, we fine-tune stochastically by sampling the squeezing and
pooling factors similar to pre-training. During validation, we use randomly selected values for $S_f$, $S_k$, and $S_q$. This model allows for a smooth
trade-off between WER and inference time for different settings of squeeze and
pooling factors used during inference. \\

\noindent\textbf{Deterministic Fine-tuning}
In this setting, we fine-tune the model for a fixed configuration of squeeze
and pooling factors. In contrast to stochastic fine-tuning, this model can only
be inferred with the selected configuration. However, we find that this gives
better WER. Note that, for each
configuration, we fine-tune the same stochastically pre-trained model resulting
in significant computational savings as pre-training typically requires more
computational resources and time.

\subsection{Evaluation}
Similar to \cite{felix2021performance}, we consider the following metrics:
pre-training time, inference time, and word error rate (WER) for model
efficiency.  We report inference times (in seconds) on \emph{dev other} split using CTC greedy decoding on NVIDIA V100
with FP32 operations.  We use the 4-gram language model (LM) and wav2letter decoder
\cite{pratap2019wav2letter} for decoding with language model (LM).
Similar to \cite{felix2021performance}, we do not tune hyper-parameters 
when decoding with LM. We use the default LM weight 2, word score -1, and beam size 50.

\subsubsection{Inference with Stochastic Models}
For our proposed stochastically pre-trained models fine-tuned in stochastic or
deterministic settings, we provide inference time and WER for various
configurations of squeeze, query and key-value pooling factors.

\section{Results}

In the following, we first analyze the performance of stochastic W2V2 model for
different query and key-value mean-pooling choices.  We then discuss the
trade-offs for the stochastically trained W2V2 model against the original W2V2
and SEW models. Finally, we present the WER results for Base and Large on the
\emph{clean} and \emph{other} parts of the Librispeech test sets.

\label{sec:results}
\subsection{How much query and key-value pooling?}

We consider (a) $\{1,2,3\}$ and (b) $\{1,2\}$ as the  two choices of 
key-value and query pooling sets to analyze the WER and inference time trade-offs for
base models pre-trained for $100$K steps. 

In Table \ref{tab:qkv-pooling}, we present the pre-training time (PT) for each of
these models. We can see that the pre-training time for both (a) and (b) is
quite similar. Both of these are faster than the \mbox{W2V2-B} model and slower than
the SEW-B model. This is expected because we occasionally sample the
squeeze and pooling factors as $1$ in which case the computation time for our
model would be higher than that of SEW-B model.

\begin{figure}
    \begin{subfigure}{0.9\columnwidth}
        \includegraphics[width=\columnwidth]{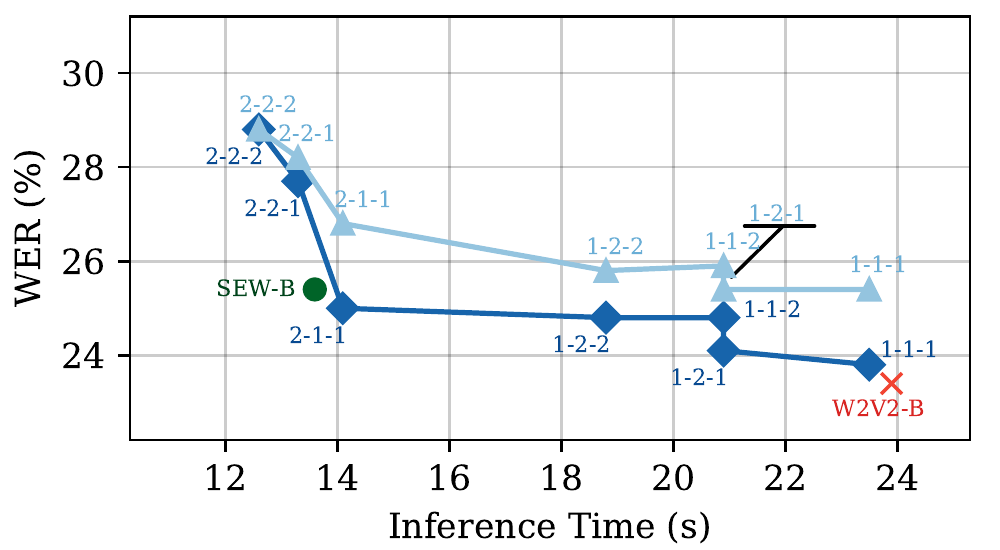}
    \end{subfigure}

        \centering
        \includegraphics[width=0.9\columnwidth]{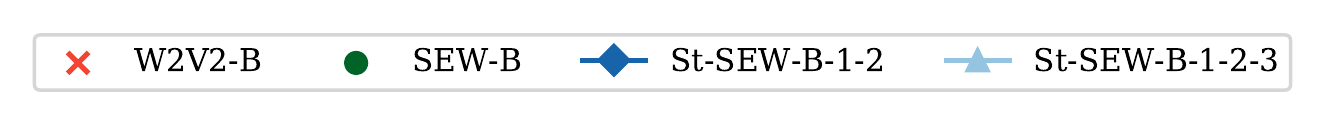}
        \caption{
            We compare the different query and key-value pooling options for
            St-SEW-B models trained for $100$K steps. The numbers near the
            datapoints denote the inference configuration in the order: squeeze ($S_f$), key-value pooling ($S_k$), and query pooling ($S_q$) factors. We can see that using pooling factors of $\{1,2\}$
            results in a better performance trade-off.
        }

    \label{fig:qkv_pooling}
\end{figure}

In Fig. \ref{fig:qkv_pooling}, we present the inference time and WER trade-offs
for different models. We plot the WER obtained on \emph{dev-other} split when
the same model is then inferred with different values for squeeze ($S_f$),
key-value ($S_k$) and query pooling ($S_q$) pooling factors indicated on the
figures as triplet in the order $S_f$-$S_k$-$S_q$ . We also train the W2V2-B
and SEW-B models for comparison.  We can see that the model trained with
pooling factors sampled from $\{1,2\}$ outperforms the model that samples pooling
factors from $\{1,2,3\}$.  We also see that the performance for this model is
similar to the non-stochastic W2V2-B and SEW-B models. In
all subsequent experiments, we always choose the query and key-value pooling
factors  from $\{1,2\}$ .

\begin{table}
  \caption{
    Comparing pre-training time for different choices of query and key-pooling
    factors for St-SEW-B models trained for $100$K steps
  }
  \centering
    \begin{tabular}{l c c}
    \toprule
    {Model}         & {\{$S_f$,$S_k$,$S_q$\}}          & {PT (hours)} \\ 
    \midrule
    W2V2-B          & {\{1\},\{1\},\{1\}}              &  29.2 \\
    SEW-B           & {\{2\},\{1\},\{1\}}              &  20.7 \\
    St-SEW-B-1-2    & {\{1,2\},\{1,2\},\{1,2\}}        &  24.9 \\
    St-SEW-B-1-2-3  & {\{1,2\},\{1,2,3\},\{1,2,3\}}    &  24.6 \\
    \bottomrule
  \end{tabular}
  \label{tab:qkv-pooling}
\end{table}

\subsection{On-demand compute reduction inference}

%
%

\begin{figure*}[ht]

    \centering
    \begin{subfigure}{0.45\textwidth}
        \includegraphics[width=\columnwidth]{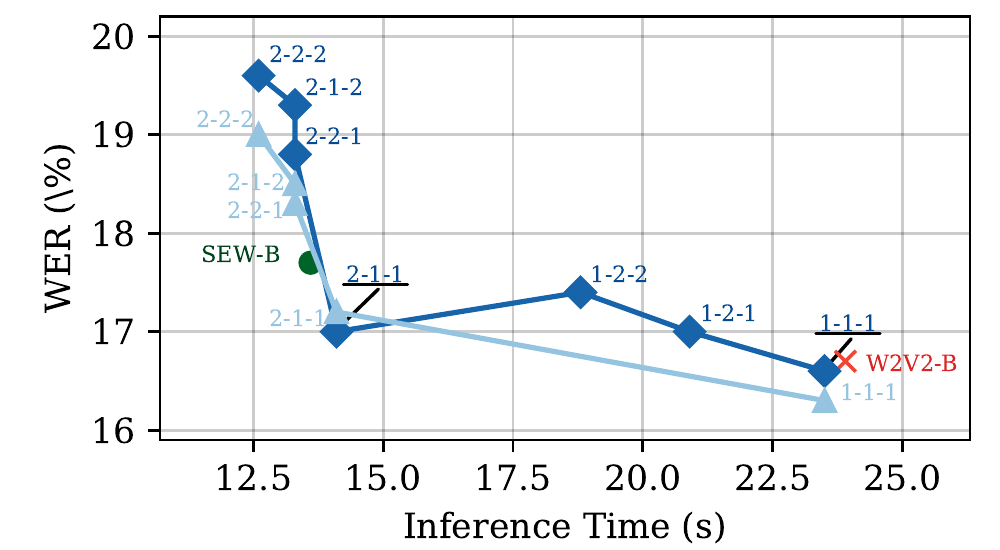}
        \caption{Base Model}
    \end{subfigure}
    \begin{subfigure}{0.45\textwidth}
        \includegraphics[width=\columnwidth]{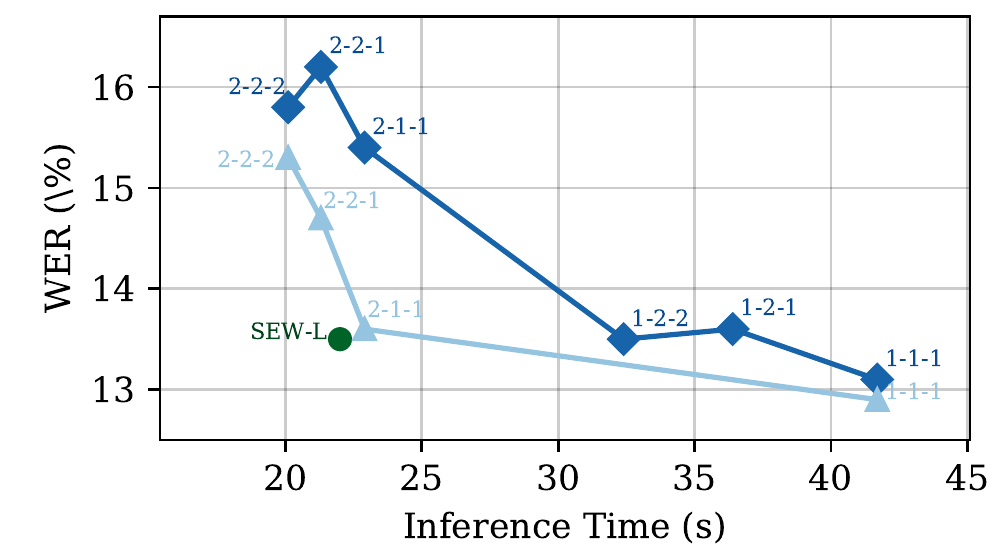}
        \caption{Large Model}
    \end{subfigure}

    \includegraphics[width=1.3\columnwidth]{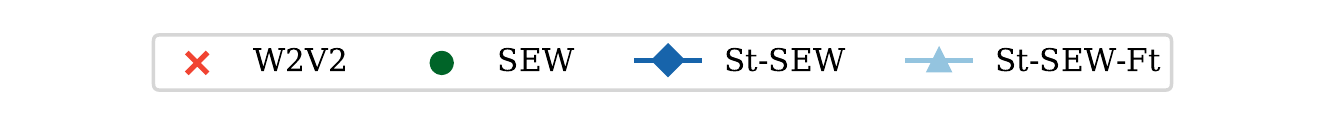}
    \caption{WER and inference time tradeoff for different Base and Large
            models. The numbers near the datapoints denote the configuration
            used during inference in the following orders:
            ($S_f$-$S_k$-$S_q$). We see that the stochastically trained model
            provides a smooth trade-off between WER and inference times.
            Fine-tuning the pre-trained to a specific configuration of interest
            improves the WER significantly.}
    \label{fig:compute-tradeoff-dev}
\end{figure*}

In this section, we compare the WER and inference time trade-offs for Base and
Large models. We pre-train each model for $400$K steps followed by fine-tuning
on the $10$h transcribed subset of Librispeech dataset. We use
\mbox{\emph{St-SEW-B-Ft}} and \mbox{\emph{St-SEW-L-Ft}} to denote the stochastic pre-trained
models fine-tuned in deterministic setting as discussed before. We evaluate the
inference time and WER for the following configurations of ($S_f$,$S_k$,$S_q$)
factors: (a) (1,1,1), (b) (2,1,1) (c) (2,2,1) (d) (2,2,2). We select (1,1,1)
and (2,1,1) to compare against the W2V2 and SEW models respectively. We then
increase the query and key pooling to further increase the overall compression
resulting in faster  inference. We skip results for the (2,1,2) as it is
very similar to (2,2,1). 

Fig. \ref{fig:compute-tradeoff-dev} shows the WER and inference-time for Base
and Large models evaluated on the \emph{dev-other} portion of the Librispeech
dataset.  For Base models, the same stochastically fine-tuned (St-SEW) model performs only marginally worse than W2V2 and SEW models when inferred using the corresponding configurations. There is a bigger performance difference in the case of Large models especially when $S_f=2$ during inference. However, the difference in performance can be 
recovered, if we fine-tune the models to specific configurations (St-SEW-L-Ft).
Depending on the application requirements, we can choose different operating points
(configurations) for an on-demand compute reduction for a smooth trade-off in
WER.

For Large models, we find that W2V2-L pre-training is unstable and
diverges quickly. In contrast to this the stochastic model pre-training and
fine-tuning is stable. We always use post-layer norm for Transformer encoder. In \cite{baevski2020wav2vec}, for stable pre-training, Large models are trained with pre-layer norm and  convolutional front-end with extra normalization layers resulting in significantly higher training time and inference latency.

\subsection{Test Set Evaluation}

\begin{table}[h]
  \caption{
      Inference times and WER result for base models on \emph{other} and \emph{clean} test sets. We report the WER obtained using greedy decoding
      as well as 4-gram LM (in parenthesis).
  }
  \label{tab:res-test-base}
  \centering
  \scalebox{0.78}{
  \begin{tabular}{l l c c c c}
    \toprule
      & & \multicolumn{4}{c}{Inference ($S_f$,$S_k$,$S_q$)} \\
    \cmidrule{3-6}
      & {Model}   & {1,1,1} & {2,1,1} & {2,2,1} & {2,2,2} \\
    \midrule
    \multicolumn{6}{c}{Inference times (dev other)} \\ 
    \midrule
      (a) & W2V2-B      & 23.9  &  -   & -    & -   \\
      (b) & SEW-B       & -     & 13.6 & -    & -   \\
      (c) & St-SEW-B    & 23.5  & 14.0 & 13.3 & 12.6 \\
    \midrule
    \multicolumn{6}{c}{Word Error Rate Results (test clean)} \\
    \midrule
      (d) & W2V2-B      & 9.5 (5.0)   &  -          & -   & - \\
      (e) & SEW-B       & -           & 10.2 (4.9)  & -   & - \\
      (f) & St-SEW-B    & 9.7 (5.1)   & 9.8 (5.2)   & 11.0 (5.4)   & 11.4 (5.4) \\
      (g) & St-SEW-B-Ft & 9.4 (5.0)   & 10.0 (4.9)  & 10.8 (5.1)   & 11.2 (5.2) \\
    \midrule
    \multicolumn{6}{c}{Word Error Rate Results (test other)} \\
    \midrule
      (h) & W2V2-B      & 16.7 (10.7)   &  -            & -             & - \\
      (i) & SEW-B       & -             & 17.6 (10.8)   & -             & - \\
      (j) & St-SEW-B    & 16.8 (11.0)  & 17.3 (11.3)    & 19.0 (11.6)   & 19.9 (11.8) \\
      (k) & St-SEW-B-Ft & 16.5 (10.8)  & 17.3 (10.8)    & 18.5 (11.3)   & 19.3 (11.7) \\
    \bottomrule
  \end{tabular}
  }
\end{table}

We present the WER obtained by Base and Large models on the
\emph{clean} and \emph{other} portion of the Librispeech test sets. From Table
\ref{tab:res-test-base}, we see that similar to \emph{dev other}, the WER for \mbox{\emph{St-SEW-B}} model is very close to W2V2 and SEW models in the corresponding configurations. We also see that fine-tuning to specific configurations (\emph{St-SEW-B-Ft}) gives slightly better WER.

Table \ref{tab:res-test-large} presents the results for Large models where we find that \emph{St-SEW-L} does not perform as well when the squeeze factor is set to $2$. We suspect that the randomly selected values for $S_f$, $S_k$, and $S_q$ during validation may cause the selected model to be better for some configurations than other. However, we again see that for \emph{St-SEW-L-Ft} models, the performance is very close to that of the \emph{SEW-L} model.

\begin{table}[h]
  \caption{
      Inference times and WER result for large models on \emph{other} and \emph{clean} test sets. We report the WER obtained using greedy decoding
      as well as 4-gram LM (in parenthesis).
  }
  \label{tab:res-test-large}
  \scalebox{0.78}{
  \centering
  \begin{tabular}{l l c c c c}
    \toprule
      & & \multicolumn{4}{c}{Inference ($S_f$,$S_k$,$S_q$)} \\
    \cmidrule{3-6}
      & {Model}         & {1,1,1} & {2,1,1} & {2,2,1} & {2,2,2} \\
    \midrule
    \multicolumn{6}{c}{Inference times (dev other)} \\ 
    \midrule
      (a) & SEW-L       & -     & 22.0 & -    & -   \\
      (b) & St-SEW-L    & 41.7  & 22.9 & 21.3 & 20.1 \\
    \midrule
    \multicolumn{6}{c}{Word Error Rate Results (test clean)} \\
    \midrule
      (c) & SEW-L       & -         & 8.8  (4.6)  &  -   & -    \\
      (b) & St-SEW-L    & 8.5 (4.8) & 10.2 (5.1)  &  10.4 (5.4) & 9.8 (5.1) \\
      (c) & St-SEW-L-Ft & 8.4 (4.7) & 8.9 (4.5)   &  9.6 (4.7)  & 9.8 (4.8) \\
    \midrule
    \multicolumn{6}{c}{Word Error Rate Results (test other)} \\
    \midrule
      (d) & SEW-L       & -          & 13.9 (9.1)   &  -   & -    \\
      (e) & St-SEW-L    & 13.6 (9.3) & 16.2 (10.1)  & 16.6 (10.1)   & 16.3 (10.2)    \\
      (f) & St-SEW-L-Ft & 13.4 (9.1) & 14.1 (9.1)   & 15.2 (9.5)    & 15.7 (9.9) \\
    \bottomrule
  \end{tabular}
  }
\end{table}

\section{Conclusion}
\label{sec:conclusions}
We proposed a stochastic compression technique for compute reduction during
W2V2 pre-training as well as for on-demand compute reduction during inference. We show
that stochastically pre-trained and fine-tuned models  provide multiple
operating points with smooth performance trade-off for different applications.
We further show that fine-tuning the stochastically pre-trained model to a
specific configuration provides the same WER as a model pre-trained and
fine-tuned from scratch for the same configuration.

Currently, we use the same squeeze and pooling factors for all utterances. In
future, we will explore techniques to adaptively choose the compression factors
depending on the input utterance to improve the WER and inference time
trade-off.

\bibliographystyle{IEEEtran}
\bibliography{refs}


\end{document}